\newif\ifanonymousversion
\anonymousversionfalse

\documentclass[sigconf]{acmart}

\setcopyright{acmcopyright}
\renewcommand\footnotetextcopyrightpermission[1]{} 

\usepackage{graphicx}
\usepackage{xcolor}
\usepackage{soul}
\usepackage{caption}
\usepackage{subcaption}
\usepackage{multirow}
\usepackage{multicol}
\usepackage{adjustbox}
\usepackage{hyperref}

\newcommand{\nsf}[1]{\href{https://www.nsf.gov/awardsearch/showAward?AWD_ID=#1}{#1}}

\begin{document}

\title[{Snowball Adversarial Attack on Traffic Sign Classification}]{{Snowball Adversarial Attack on Traffic Sign Classification}\vspace{1cm}}

\ifanonymousversion
\author{Anonymous Submission}

\else

\author{Anthony Etim}
\affiliation{%
  \institution{Yale University}
  \city{New Haven} 
  \state{CT} 
  \country{USA}
}
\email{anthony.etim@yale.edu}

\author{Jakub Szefer}
\affiliation{%
  \institution{Northwestern University}
  \city{Evanston} 
  \state{IL} 
  \country{USA}
}
\email{jakub.szefer@northwestern.edu}

\fi

\begin{abstract}
    Adversarial attacks on machine learning models often rely on small, imperceptible perturbations to mislead classifiers. Such strategy focuses on minimizing the visual perturbation for humans so they are not confused, and also maximizing the misclassification for machine learning algorithms. An orthogonal strategy for adversarial attacks is to create perturbations that are clearly visible but do not confuse humans, yet still maximize misclassification for machine learning algorithms. This work follows the later strategy, and demonstrates instance of it through the Snowball Adversarial Attack in the context of traffic sign recognition. The attack leverages the human brain's superior ability to recognize objects despite various occlusions, while machine learning algorithms are easily confused. The evaluation shows that the Snowball Adversarial Attack is robust across various images and is able to confuse state-of-the-art traffic sign recognition algorithm. The findings reveal that Snowball Adversarial Attack can significantly degrade model performance with minimal effort, raising important concerns about the vulnerabilities of deep neural networks and highlighting the necessity for improved defenses for image recognition machine learning models.
\end{abstract}

\maketitle

\pagestyle{plain}

\section{Introduction}
\label{sec_introduction}

Deep neural networks (DNNs) have become integral to modern computer vision applications, particularly in autonomous driving, where traffic sign recognition plays a crucial role in decision-making. However, these models remain highly susceptible to adversarial attacks, where subtle, carefully placed perturbations can mislead classification systems~\cite{szegedy2013intriguing, goodfellow2014explaining}. Such attack strategies focus on minimizing the visual perturbation for humans so they are not confused while maximizing the misclassification for machine learning algorithms.

In an orthogonal attack strategy, growing number of adversarial image attacks focuses on exploiting image perturbations that are visible, but not confusing, to humans, while they make machine learning algorithms to fail to correctly recognize images. In particular, researchers have focused on naturally occurring artifacts such as leaves, shadows, and lighting variations that can be exploited to create adversarial perturbations to images, subtly altering a model’s perception without drawing human suspicion~\cite{zhong2022shadows, hsiao2024natural,etim2024fallleafadversarialattack}. These physical attacks are particularly concerning because they can also be executed in natural environments by physically modifying the objects being captured with digital cameras, without requiring direct access to a machine learning model or digital input data.

These attacks highlight the fragility of deep learning models in uncontrolled environments, where even minor environmental changes can cause misclassification. Building upon this idea, we introduce the Snowball Adversarial Attack, a novel physical attack that simulates snow accumulation on traffic signs to induce misclassification. This attack leverages the idea that naturally occurring snowfall or snowballs stuck to a traffic sign are naturally ignored by drivers, yet they can correctly recognize the street signs -- but machine learning models are confused. Unlike traditional adversarial attacks that require small, precisely crafted perturbations, the Snowball Adversarial Attack utilizes a progressive occlusion strategy, where obvious patches of snow are placed at strategic locations over street signs. A search algorithm is utilized to find the best location and is shown to be effective for many street sign types and various snowballs or snow patches.

This attack is particularly dangerous in real-world winter conditions, where snowfall already introduces visual obstructions and reduces the contrast of traffic signs against their backgrounds. Because autonomous vehicles and traffic monitoring systems rely heavily on visual perception, an attacker can exploit this vulnerability by carefully various amounts of snow in key regions of a sign, disrupting classification while maintaining a seemingly natural appearance. This makes the Snowball Adversarial Attack both stealthy and highly effective, as it blends seamlessly into common winter weather conditions, making it difficult to detect even through human oversight.

In addition to presenting the new attack, a secondary contribution of this work is a digital framework used to evaluate the attacks. It makes the evaluation more scalable, but also removes need to physically alter the street signs -- which could be impractical, dangerous or even illegal. The digital framework is used in our evaluation and analysis of the attack's effectiveness.

The Snowball Adversarial Attack poses a significant threat to autonomous vehicles and intelligent transportation systems. Since self-driving cars rely heavily on computer vision-based classifiers for real-time decision-making, an attack that misclassifies a stop sign as a speed limit sign could have severe consequences. Given the stealthy and naturally occurring nature of snow this attack may be hard to detect and there is need for creating more robust traffic sign recognition models.

\subsection{Contributions}

The contributions of this work are as follows:

\begin{enumerate}

    \item We propose the Snowball Adversarial Attack, a novel adversarial attack that leverages snow-like obstructions to manipulate traffic sign classification.
    
    \item We demonstrate that strategic snow placement on specific regions of traffic signs maximizes misclassification rates across different sign types, showing a universal effect across various traffic sign classes.
    
    \item We evaluate the attack’s effectiveness using our digital framework, leveraging real street signs from Street View and various realistic snowballs or patches created using generative machine learning models.

    \item We demonstrate the ability of the attack to cause misclassification of all the street signs we have tested.
    
\end{enumerate}

\section{Background}
\label{background}

In this section, we provide an overview of adversarial attacks in machine learning, focusing on their implications for traffic sign classification systems. We also introduce the LISA dataset, a widely used benchmark for U.S. traffic sign recognition, and discuss LISA-CNN, the convolutional neural network model used in this work as a victim classifier.

\subsection{Adversarial Attacks}

Adversarial attacks exploit the vulnerabilities of deep neural networks (DNNs) by introducing small, carefully crafted perturbations to input data, causing the model to misclassify objects while remaining undetectable to human observers~\cite{szegedy2013intriguing, goodfellow2014explaining}. Other attacks have explored adding larger perturbations, which are ignored by humans, but confuse machine learning models~\cite{eykholt2018robust, zhong2022shadows}.
Adversarial attacks can be broadly categorized into digital attacks, where perturbations are applied directly to input images, and physical attacks, where adversarial modifications are introduced into real-world environments. 

Among digital attacks, early attack methods, such as the Fast Gradient Sign Method (FGSM)\cite{goodfellow2014explaining} and Projected Gradient Descent (PGD)\cite{madry2017towards}, rely on perturbing individual images to fool a model. More advanced approaches, like universal adversarial perturbations (UAPs), generate a single perturbation that generalizes across multiple inputs, making attacks more practical for real-world scenarios~\cite{moosavi2017universal}.

Among physical attacks, recent studies have shown that physical adversarial attacks pose a significant challenge to vision-based models. Attacks using stickers, shadows, and light-based manipulations have successfully deceived traffic sign classifiers by subtly altering key regions of an image without requiring direct access to the model~\cite{eykholt2018robust, zhong2022shadows, hsiao2024natural}. Attacks using leaves have also been used to mislead traffic sign classifiers as well~\cite{etim2024fallleafadversarialattack}. These attacks highlight the importance of understanding adversarial vulnerabilities beyond digital spaces and accounting for real-world environmental factors that may compromise model robustness.

\begin{figure*}[th!]
    \begin{subfigure}[b]{0.19\textwidth}
        \centering
\includegraphics[width=2.2cm]{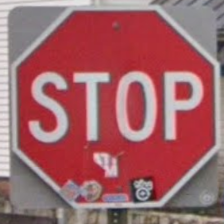}
        \caption{\footnotesize \centering Stop \newline Test Image}
        \label{fig:stop_image}
    \end{subfigure}
    \hfill
    \begin{subfigure}[b]{0.19\textwidth}
        \centering
        \includegraphics[width=2.2cm]{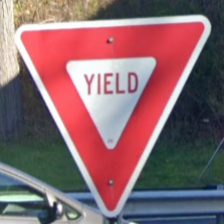}
        \caption{\footnotesize \centering Yield\newline Test Image}
    \label{fig:yield_image}
    \end{subfigure}
    \hfill
     \begin{subfigure}[b]{0.19\textwidth}
        \centering
    \includegraphics[width=2.2cm]{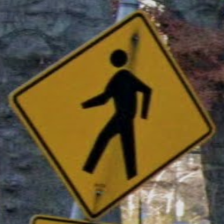}
        \caption{\footnotesize \centering Ped. Crossing\newline Test Image}
    \label{fig:pedestrian_image}
    \end{subfigure}
    \hfill
    \begin{subfigure}[b]{0.19\textwidth}
        \centering
        \includegraphics[width=2.2cm]{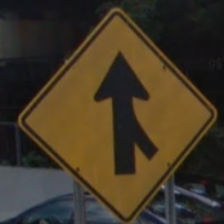}
        \caption{\footnotesize \centering Merge\newline Test Image}
    \label{fig:merge_image}
    \end{subfigure}
    \hfill
    \begin{subfigure}[b]{0.19\textwidth}
        \centering
        \includegraphics[width=2.2cm]{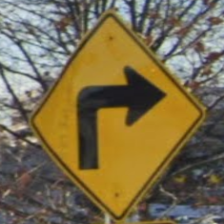}
        \caption{\footnotesize \centering Turn Right\newline Test Image}
    \label{fig:right_image}
    \end{subfigure}
    \hfill
    \caption{Street View test images used in evaluation of the attacks.}
    \label{fig:Test_Images}
\end{figure*}

\begin{figure*}[th!]
    \begin{subfigure}[b]{0.19\textwidth}
        \centering
\includegraphics[width=2.2cm]{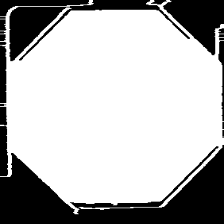}
        \caption{\footnotesize \centering Stop\newline Output Mask}
        \label{fig:stop_output_mask}
    \end{subfigure}
    \hfill
    \begin{subfigure}[b]{0.19\textwidth}
        \centering
        \includegraphics[width=2.2cm]{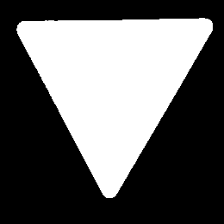}
        \caption{\footnotesize \centering Yield\newline Output Mask}
    \label{fig:yield_output_mask}
    \end{subfigure}
    \hfill
    \begin{subfigure}[b]{0.19\textwidth}
        \centering
    \includegraphics[width=2.2cm]{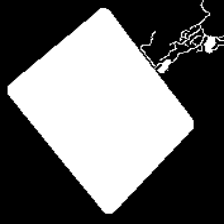}
        \caption{\footnotesize \centering Ped. Crossing\newline Output Mask}
    \label{fig:pedestrian_output_mask}
    \end{subfigure}
    \hfill
    \begin{subfigure}[b]{0.19\textwidth}
        \centering
        \includegraphics[width=2.2cm]{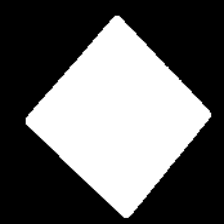}
        \caption{\footnotesize \centering Merge\newline Output Mask}
    \label{fig:merge_output_mask}
    \end{subfigure}
    \hfill
    \begin{subfigure}[b]{0.19\textwidth}
        \centering
        \includegraphics[width=2.2cm]{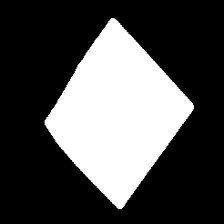}
        \caption{\footnotesize \centering Turn Right\newline Output Mask}
    \label{fig:right_output_mask}
    \end{subfigure}
    \hfill
    \caption{Street View test image masks used in placing snowball patches.}
     \label{fig:Output Masks}
\end{figure*}

\begin{figure*}[th!]
    \centering
    \begin{subfigure}[b]{0.1\textwidth}
        \centering
        \includegraphics[width=\linewidth]{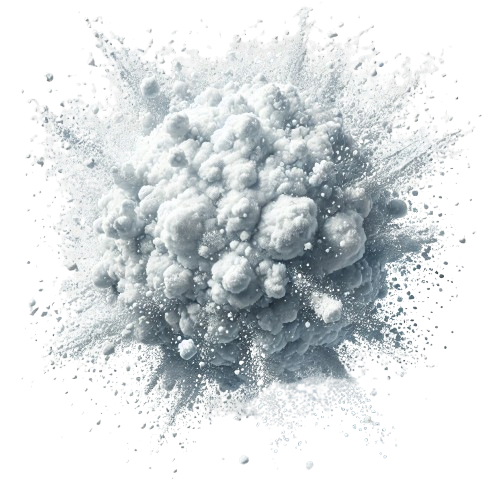}
        \caption{\footnotesize \centering Snowball 1}
        \label{fig:snowball_1}
    \end{subfigure}
    \begin{subfigure}[b]{0.1\textwidth}
        \centering
        \includegraphics[width=\linewidth]{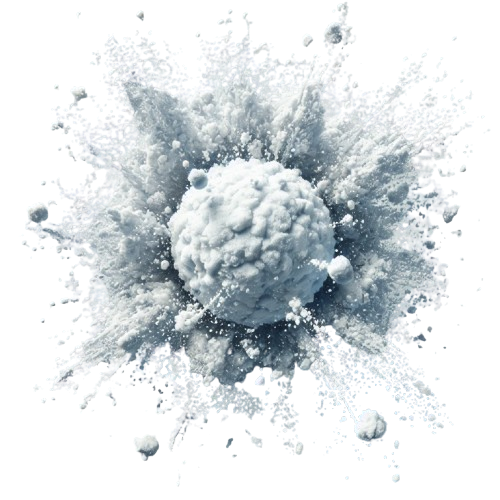}
        \caption{\footnotesize \centering Snowball 2}
        \label{fig:snowball_2}
    \end{subfigure}
    \begin{subfigure}[b]{0.1\textwidth}
        \centering
        \includegraphics[width=\linewidth]{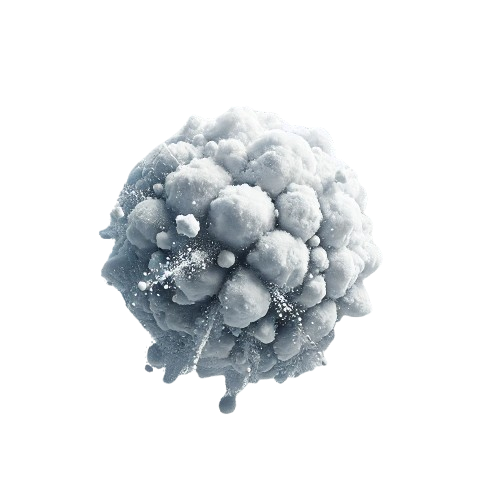}
        \caption{\footnotesize \centering Snowball 3}
        \label{fig:snowball_3}
    \end{subfigure}
    \begin{subfigure}[b]{0.1\textwidth}
        \centering
       \includegraphics[width=\linewidth]{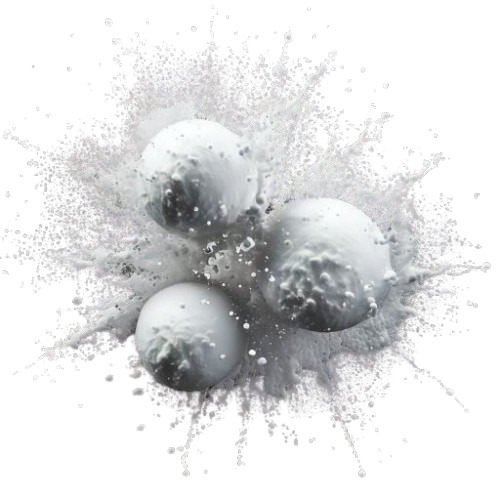}
        \caption{\footnotesize \centering Snowball 4}
        \label{fig:snowball_4}
    \end{subfigure}
    \begin{subfigure}[b]{0.1\textwidth}
        \centering
        \includegraphics[width=\linewidth]{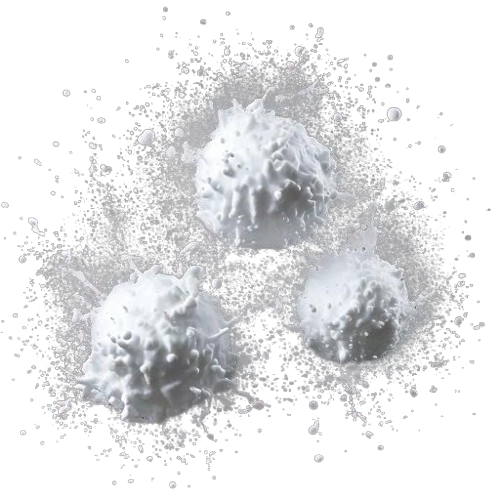}
        \caption{\footnotesize \centering Snowball 5}
        \label{fig:snowball_5}
    \end{subfigure}
    \begin{subfigure}[b]{0.1\textwidth}
        \centering
        \includegraphics[width=\linewidth]{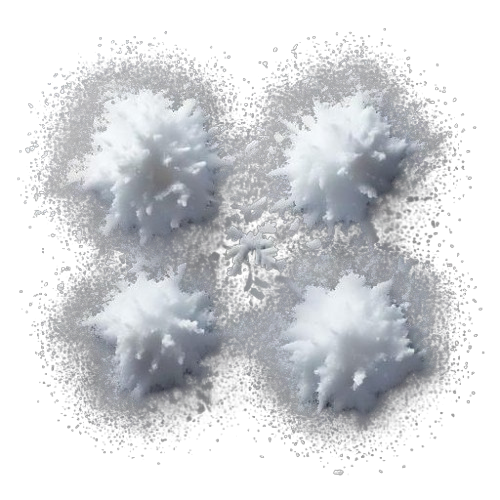}
        \caption{\footnotesize \centering Snowball 6}
        \label{fig:snowball_6}
    \end{subfigure}
    \begin{subfigure}[b]{0.1\textwidth}
        \centering
       \includegraphics[width=\linewidth]{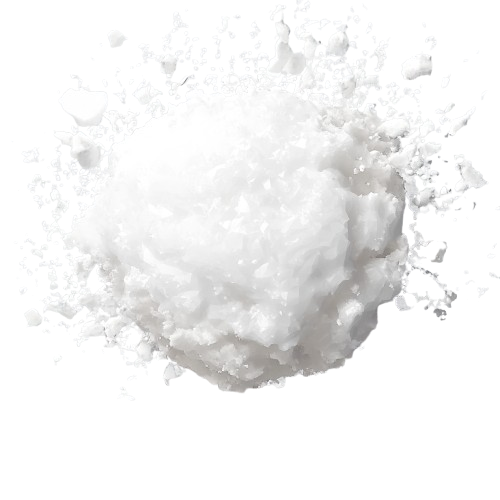}
        \caption{\footnotesize \centering Snowball 7}
        \label{fig:snowball_7}
    \end{subfigure}
    \begin{subfigure}[b]{0.1\textwidth}
        \centering
       \includegraphics[width=\linewidth]{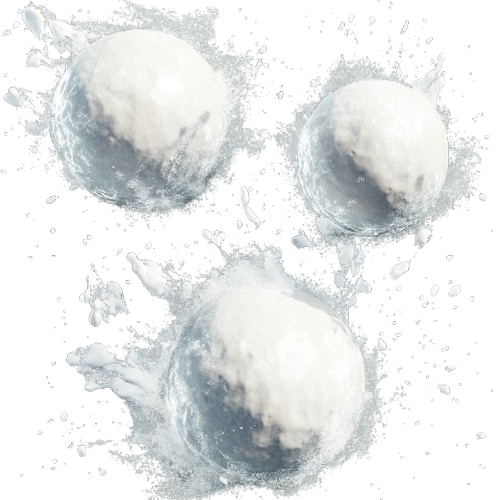}
        \caption{\footnotesize \centering Snowball 8}
        \label{fig:snowball_8}
    \end{subfigure}
    \begin{subfigure}[b]{0.1\textwidth}
        \centering
       \includegraphics[width=\linewidth]{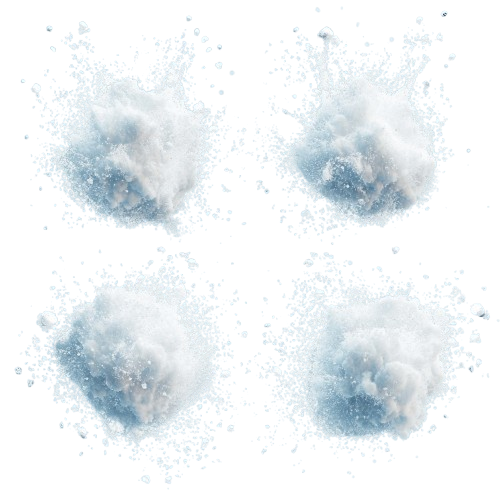}
        \caption{\footnotesize \centering Snowball 9}
        \label{fig:snowball_9}
    \end{subfigure}
\caption{Generated snowball patches used for testing.}
\label{fig:snowball_adversarial_image}
\end{figure*}

\subsection{LISA Dataset}

The LISA (Laboratory for Intelligent and Safe Automobiles) dataset is a widely used benchmark for traffic sign classification, containing images of 47 different U.S. traffic signs collected under varying environmental conditions, lighting, and viewpoints~\cite{lisa}. However, due to class imbalances, prior research has often focused on the 16 most frequently occurring traffic signs, improving model performance by ensuring sufficient training data for each class~\cite{hsiao2024natural}. The dataset is commonly used to evaluate the robustness of machine learning models deployed in various autonomous driving systems.

\subsection{Street View Street Signs}

This work evaluates attacks using images sourced from Google Street View~\cite{googleExploreStreet}, which provides real-world depictions of street signs in various environments. By modifying these images, we can assess the effectiveness of attacks in a setting that closely resembles real-world conditions without physically altering street signs. This approach ensures that evaluations remain safe, ethical, and reflective of real-world adversarial scenarios.

Additionally, it is important to note that the LISA-CNN model was not trained on Street View images. Instead, it was trained on the LISA dataset, which consists of traffic sign images collected in controlled settings. The use of Street View images for testing ensures a clear separation between training and evaluation data, reducing the risk of overfitting and making the results more representative of real-world scenarios.

\subsection{LISA-CNN}

LISA-CNN is a convolutional neural network designed for traffic sign classification using the LISA dataset. The architecture consists of three convolutional layers followed by fully connected layers, trained to recognize traffic signs under diverse real-world conditions~\cite{eykholt2018robust}. The model has demonstrated high accuracy in detecting and classifying traffic signs, making it a suitable victim model for studying adversarial vulnerabilities.

In this work, we employ LISA-CNN to evaluate the effectiveness of the Snowball Adversarial Attack. The model processes real-world traffic sign images, allowing us to assess how snow-based perturbations impact recognition accuracy. By systematically analyzing the attack’s effectiveness, we provide insights into how naturally occurring occlusions, such as snowfall, can degrade the performance of machine learning models used in self-driving cars and other autonomous driving~systems.

\section{Threat Model and Assumptions}

In this work, we consider an adversary who seeks to manipulate a deep learning-based traffic sign classification system by leveraging snow accumulation as a natural physical adversarial perturbation. The adversary exploits the fact that machine learning models heavily rely on visual features, making them vulnerable to occlusions that alter key sign characteristics. By strategically placing snow-like obstructions on traffic signs, the adversary aims to induce misclassification, leading to potential safety risks in autonomous driving systems.

The adversary operates under a black-box setting, meaning they do not have access to the model’s architecture, parameters, or training data. The adversary has physical access to traffic signs and can either rely on naturally accumulating snow or manually apply snow patches at strategic locations on the signs. The attack is particularly dangerous in winter conditions, where snow accumulation is common, making it difficult to distinguish between intentional and unintentional obstructions.

Despite its practicality, the attack is subject to several constraints. Natural weather conditions, such as wind or melting snow, may reduce its effectiveness over time. Additionally, excessive occlusion may be noticeable to human drivers, potentially limiting its impact in mixed human-autonomous driving environments.

\section{Snowball Adversarial Attack on Street Signs}

This section introduces the Snowball Adversarial Attack, an adversarial strategy that incrementally amplifies perturbations to mislead traffic sign classification systems. Inspired by the accumulation of snow or snowball patch on a street sign, the attack progressively builds upon small, natural-looking modifications -- such as occlusions from snow -- until misclassification occurs. Tested on real-world traffic sign images from Street View, the Snowball Adversarial Attack demonstrates its effectiveness against robust recognition models, highlighting the growing threat of evolving adversarial manipulations.

\subsection{Street Signs}

Street signs serve as essential components of traffic regulation and road safety, providing critical information for both human drivers and autonomous systems. However, machine learning-based traffic sign recognition models remain susceptible to adversarial attacks, where minor perturbations—such as occlusions from snow, dirt, or natural wear—can significantly impact classification accuracy. The Snowball Adversarial Attack exploits this vulnerability by gradually accumulating adversarial perturbations, simulating realistic environmental changes that degrade model performance over time. To evaluate the effectiveness of this attack, we utilize real Street View images~\cite{googleExploreStreet}, enabling the simulation of real-world adversarial scenarios without the need for physical modifications to street signs. This methodology ensures a practical yet ethical assessment of the attack’s impact, demonstrating the potential risks posed to autonomous driving systems in uncontrolled environments. Figure~\ref{fig:Test_Images} illustrates  the sample street sign images.

\subsection{Street Sign Masks}

To evaluate the effectiveness of the Snowball Attack, we generated custom binary masks that precisely control the placement of accumulated perturbations on traffic signs. Our approach ensures that each perturbation aligns naturally with boundaries of the street signs and looks like natural snow, to maintain visual plausibility.

The mask generation process begins by processing input images to identify mask regions. We first convert the image to grayscale and apply Gaussian blur~\cite{hummel1987deblurring} to reduce noise. Next, Canny edge detection~\cite{rong2014improved} highlights key contours, which are refined using morphological operations to create a continuous outline. The largest contour is selected to generate a binary mask, which defines the occlusion area on the traffic sign. This method allows for precise and adaptive placement of perturbations, ensuring that occlusions appear natural under varying angles and lighting conditions. Figure~\ref{fig:Output Masks} highlights the various  masks used in the experiments.

\subsection{Generated Snowballs}

To create a diverse and representative dataset of snowball images, we utilized three commonly available image generation tools: DALL-E3~\cite{dalle3}, Adobe Firefly~\cite{adobefirefly} and Midjourney~\cite{midjourney}. Each of these platforms employs distinct generative models and rendering techniques, resulting in variations in texture, lighting, shading, and overall realism. By generating three unique snowball images from each tool, we aimed to capture a range of artistic interpretations and stylistic nuances inherent to each model. Figure~\ref{fig:snowball_adversarial_image} shows the generated snowball images.

DALL-E3, developed by OpenAI, is known for its structured and text-conditioned image synthesis, often producing images with well-defined edges and fine-grained details. Adobe Firefly, with its emphasis on creative control and photorealism, introduces additional refinements, particularly in the handling of lighting and material textures. Midjourney, on the other hand, leverages its unique diffusion-based approach to create highly stylized and visually rich compositions. The resulting images from these tools provide a basis for evaluating the consistency, diversity, and fidelity of AI-generated snowball representations.

By leveraging multiple generation sources, we ensure that our dataset includes a variety of visual perspectives, helping us assess the strengths and limitations of each AI model in replicating realistic snow formations. These images serve as the foundation for further qualitative and quantitative analysis in subsequent sections.

\subsection{Search Algorithm for Snowball Placement}

\begin{figure*}[th!]
    \centering
    \begin{subfigure}[b]{0.1\textwidth}
        \centering
        \includegraphics[width=\linewidth]{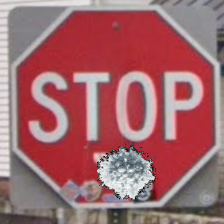}
        \caption{\footnotesize \centering Stop\newline Snowball 1}
        \label{fig:stop_snowball_1}
    \end{subfigure}
    \begin{subfigure}[b]{0.1\textwidth}
        \centering
        \includegraphics[width=\linewidth]{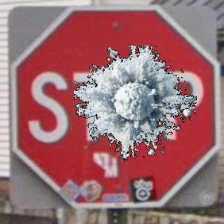}
        \caption{\footnotesize \centering Stop\newline Snowball 2}
        \label{fig:stop_snowball_2}
    \end{subfigure}
    \begin{subfigure}[b]{0.1\textwidth}
        \centering
        \includegraphics[width=\linewidth]{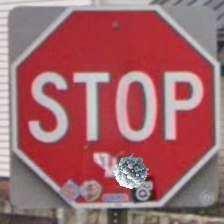}
        \caption{\footnotesize \centering Stop\newline Snowball 3}
        \label{fig:stop_snowball_3}
    \end{subfigure}
    \begin{subfigure}[b]{0.1\textwidth}
        \centering
       \includegraphics[width=\linewidth]{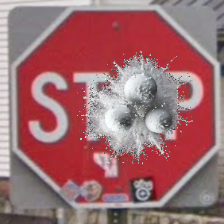}
        \caption{\footnotesize \centering Stop\newline Snowball 4}
        \label{fig:stop_snowball_4}
    \end{subfigure}
    \begin{subfigure}[b]{0.1\textwidth}
        \centering
        \includegraphics[width=\linewidth]{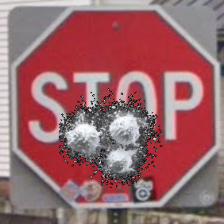}
        \caption{\footnotesize \centering Stop\newline Snowball 5}
        \label{fig:stop_snowball_5}
    \end{subfigure}
    \begin{subfigure}[b]{0.1\textwidth}
        \centering
        \includegraphics[width=\linewidth]{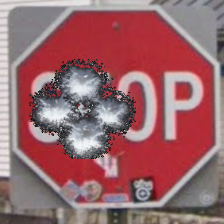}
        \caption{\footnotesize \centering Stop\newline Snowball 6}
        \label{fig:stop_snowball_6}
    \end{subfigure}
    \begin{subfigure}[b]{0.1\textwidth}
        \centering
       \includegraphics[width=\linewidth]{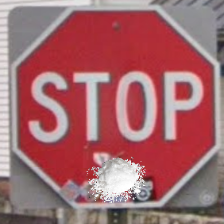}
        \caption{\footnotesize \centering Stop\newline Snowball 7}
        \label{fig:stop_snowball_7}
    \end{subfigure}
    \begin{subfigure}[b]{0.1\textwidth}
        \centering
       \includegraphics[width=\linewidth]{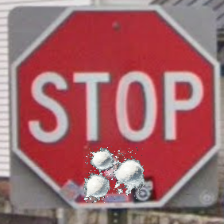}
        \caption{\footnotesize \centering Stop\newline Snowball 8}
        \label{fig:stop_snowball_8}
    \end{subfigure}
    \begin{subfigure}[b]{0.1\textwidth}
        \centering
        \includegraphics[width=\linewidth]{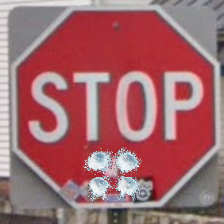}
        \caption{\footnotesize \centering Stop\newline Snowball 9}
        \label{fig:stop_snowball_9}
    \end{subfigure}
\caption{ Stop Snowball Adversarial images.}
    \label{fig:stop_snowball_adversarial_image}
\end{figure*}

\begin{figure*}[th!]
    \centering
    \begin{subfigure}[b]{0.1\textwidth}
        \centering
       \includegraphics[width=\linewidth]{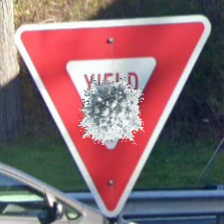}
        \caption{\footnotesize \centering Yield\newline Snowball 1}
        \label{fig:yield_snowball_1}
    \end{subfigure}
    \begin{subfigure}[b]{0.1\textwidth}
        \centering
        \includegraphics[width=\linewidth]{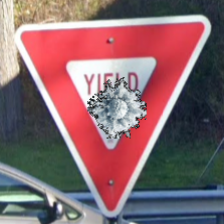}
        \caption{\footnotesize \centering Yield\newline Snowball 2}
        \label{fig:yield_snowball_2}
    \end{subfigure}
    \begin{subfigure}[b]{0.1\textwidth}
        \centering
        \includegraphics[width=\linewidth]{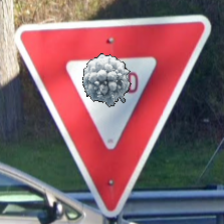}
        \caption{\footnotesize \centering Yield\newline Snowball 3}
        \label{fig:yield_snowball_3}
    \end{subfigure}
    \begin{subfigure}[b]{0.1\textwidth}
        \centering
        \includegraphics[width=\linewidth]{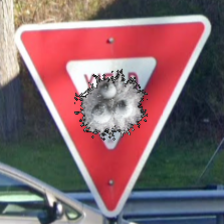}
        \caption{\footnotesize \centering Yield\newline Snowball 4}
        \label{fig:yield_snowball_4}
    \end{subfigure}
    \begin{subfigure}[b]{0.1\textwidth}
        \centering
        \includegraphics[width=\linewidth]{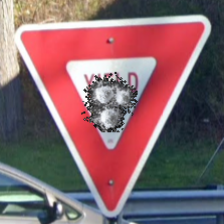}
        \caption{\footnotesize \centering Yield\newline Snowball 5}
        \label{fig:yield_snowball_5}
    \end{subfigure}
    \begin{subfigure}[b]{0.1\textwidth}
        \centering
        \includegraphics[width=\linewidth]{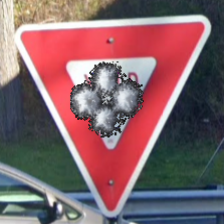}
        \caption{\footnotesize \centering Yield\newline Snowball 6}
        \label{fig:yield_snowball_6}
    \end{subfigure}
    \begin{subfigure}[b]{0.1\textwidth}
        \centering
        \includegraphics[width=\linewidth]{plots/base_images_resized/yield.png}
        \caption{\footnotesize \centering Yield\newline Snowball 7}
        \label{fig:yield_snowball_7}
    \end{subfigure}
    \begin{subfigure}[b]{0.1\textwidth}
        \centering
        \includegraphics[width=\linewidth]{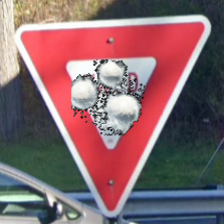}
        \caption{\footnotesize \centering Yield\newline Snowball 8}
        \label{fig:yield_snowball_8}
    \end{subfigure}
    \begin{subfigure}[b]{0.1\textwidth}
        \centering
        \includegraphics[width=\linewidth]{plots/base_images_resized/yield.png}
        \caption{\footnotesize \centering Yield\newline Snowball 9}
        \label{fig:yield_snowball_9}
    \end{subfigure}
\caption{ Yield Snowball Adversarial images. Note no effective attack was found for signs (g) and (i).}
    \label{fig:yield_snowball_adversarial_image}
\end{figure*}

\begin{figure*}[th!]
    \centering
    \begin{subfigure}[b]{0.1\textwidth}
        \centering
        \includegraphics[width=\linewidth]{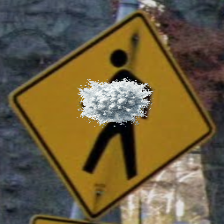}
        \caption{\footnotesize \centering Pedestrian\newline Snowball 1}
        \label{fig:pedestrian_snowball_1}
    \end{subfigure}
    \begin{subfigure}[b]{0.1\textwidth}
        \centering
        \includegraphics[width=\linewidth]{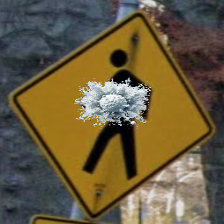}
        \caption{\footnotesize \centering Pedestrian\newline Snowball 2}
        \label{fig:pedestrian_snowball_2}
    \end{subfigure}
    \begin{subfigure}[b]{0.1\textwidth}
        \centering
        \includegraphics[width=\linewidth]{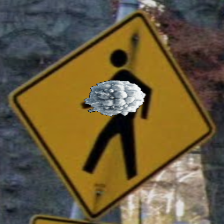}
        \caption{\footnotesize \centering Pedestrian\newline Snowball 3}
        \label{fig:pedestrian_snowball_3}
    \end{subfigure}
    \begin{subfigure}[b]{0.1\textwidth}
        \centering
        \includegraphics[width=\linewidth]{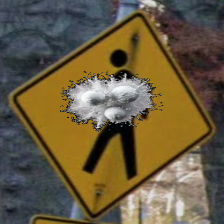}
        \caption{\footnotesize \centering Pedestrian\newline Snowball 4}
        \label{fig:pedestrian_snowball_4}
    \end{subfigure}
    \begin{subfigure}[b]{0.1\textwidth}
        \centering
        \includegraphics[width=\linewidth]{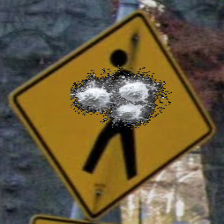}
        \caption{\footnotesize \centering Pedestrian\newline Snowball 5}
        \label{fig:pedestrian_snowball_5}
    \end{subfigure}
    \begin{subfigure}[b]{0.1\textwidth}
        \centering
        \includegraphics[width=\linewidth]{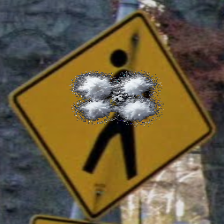}
        \caption{\footnotesize \centering Pedestrian\newline Snowball 6}
        \label{fig:pedestrian_snowball_6}
    \end{subfigure}
    \begin{subfigure}[b]{0.1\textwidth}
        \centering
        \includegraphics[width=\linewidth]{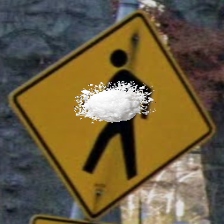}
        \caption{\footnotesize \centering Pedestrian\newline Snowball 7}
        \label{fig:pedestrian_snowball_7}
    \end{subfigure}
    \begin{subfigure}[b]{0.1\textwidth}
        \centering
        \includegraphics[width=\linewidth]{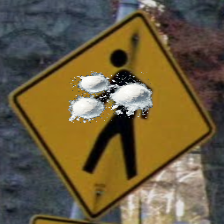}
        \caption{\footnotesize \centering Pedestrian\newline Snowball 8}
        \label{fig:pedestrian_snowball_8}
    \end{subfigure}
    \begin{subfigure}[b]{0.1\textwidth}
        \centering
        \includegraphics[width=\linewidth]{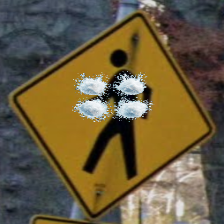}
        \caption{\footnotesize \centering Pedestrian\newline Snowball 9}
        \label{fig:pedestrian_snowball_9}
    \end{subfigure}
\caption{ Pedestrian Snowball Adversarial images.}
    \label{fig:pedestrian_snowball_adversarial_image}
\end{figure*}

\begin{figure*}[th!]
    \centering
    \begin{subfigure}[b]{0.1\textwidth}
        \centering
        \includegraphics[width=\linewidth]{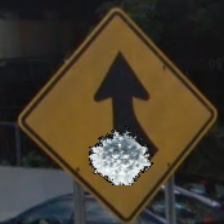}
        \caption{\footnotesize \centering Merge\newline Snowball 1}
        \label{fig:merge_snowball_1}
    \end{subfigure}
    \begin{subfigure}[b]{0.1\textwidth}
        \centering
        \includegraphics[width=\linewidth]{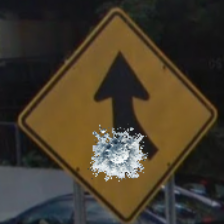}
        \caption{\footnotesize \centering Merge\newline Snowball 2}
        \label{fig:merge_snowball_2}
    \end{subfigure}
    \begin{subfigure}[b]{0.1\textwidth}
        \centering
        \includegraphics[width=\linewidth]{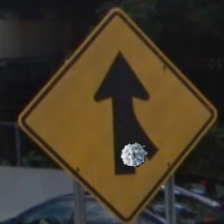}
        \caption{\footnotesize \centering Merge\newline Snowball 3}
        \label{fig:merge_snowball_3}
    \end{subfigure}
    \begin{subfigure}[b]{0.1\textwidth}
        \centering
        \includegraphics[width=\linewidth]{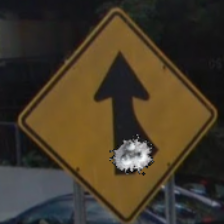}
        \caption{\footnotesize \centering Merge\newline Snowball 4}
        \label{fig:merge_snowball_4}
    \end{subfigure}
    \begin{subfigure}[b]{0.1\textwidth}
        \centering
        \includegraphics[width=\linewidth]{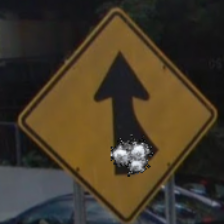}
        \caption{\footnotesize \centering Merge\newline Snowball 5}
        \label{fig:merge_snowball_5}
    \end{subfigure}
    \begin{subfigure}[b]{0.1\textwidth}
        \centering
        \includegraphics[width=\linewidth]{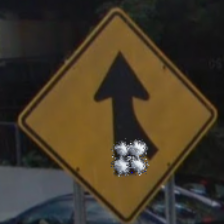}
        \caption{\footnotesize \centering Merge\newline Snowball 6}
        \label{fig:merge_snowball_6}
    \end{subfigure}
    \begin{subfigure}[b]{0.1\textwidth}
        \centering
        \includegraphics[width=\linewidth]{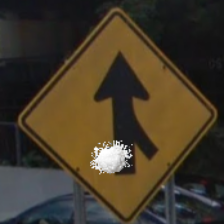}
        \caption{\footnotesize \centering Merge\newline Snowball 7}
        \label{fig:merge_snowball_7}
    \end{subfigure}
    \begin{subfigure}[b]{0.1\textwidth}
        \centering
        \includegraphics[width=\linewidth]{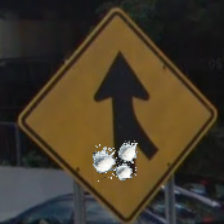}
        \caption{\footnotesize \centering Merge\newline Snowball 8}
        \label{fig:merge_snowball_8}
    \end{subfigure}
    \begin{subfigure}[b]{0.1\textwidth}
        \centering
        \includegraphics[width=\linewidth]{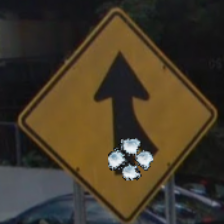}
        \caption{\footnotesize \centering Merge\newline Snowball 9}
        \label{fig:merge_snowball_9}
    \end{subfigure}
\caption{ Merge Snowball Adversarial images.}
    \label{fig:merge_snowball_adversarial_image}
\end{figure*}

\begin{figure*}[th!]
    \centering
    \begin{subfigure}[b]{0.1\textwidth}
        \centering
        \includegraphics[width=\linewidth]{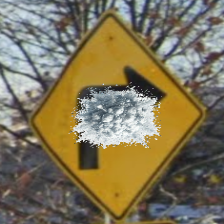}
        \caption{\footnotesize \centering Turn Right\newline Snowball 1}
        \label{fig:right_snowball_1}
    \end{subfigure}
    \begin{subfigure}[b]{0.1\textwidth}
        \centering
        \includegraphics[width=\linewidth]{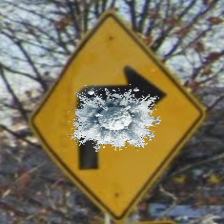}
        \caption{\footnotesize \centering Turn Right\newline Snowball 2}
        \label{fig:right_snowball_2}
    \end{subfigure}
    \begin{subfigure}[b]{0.1\textwidth}
        \centering
        \includegraphics[width=\linewidth]{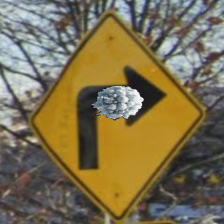}
        \caption{\footnotesize \centering Turn Right\newline Snowball 3}
        \label{fig:right_snowball_3}
    \end{subfigure}
    \begin{subfigure}[b]{0.1\textwidth}
        \centering
        \includegraphics[width=\linewidth]{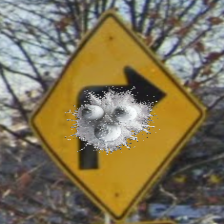}
        \caption{\footnotesize \centering Turn Right\newline Snowball 4}
        \label{fig:right_snowball_4}
    \end{subfigure}
    \begin{subfigure}[b]{0.1\textwidth}
        \centering
        \includegraphics[width=\linewidth]{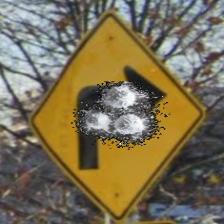}
        \caption{\footnotesize \centering Turn Right\newline Snowball 5}
        \label{fig:right_snowball_5}
    \end{subfigure}
    \begin{subfigure}[b]{0.1\textwidth}
        \centering
        \includegraphics[width=\linewidth]{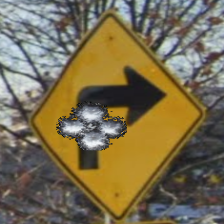}
        \caption{\footnotesize \centering Turn Right\newline Snowball 6}
        \label{fig:right_snowball_6}
    \end{subfigure}
    \begin{subfigure}[b]{0.1\textwidth}
        \centering
        \includegraphics[width=\linewidth]{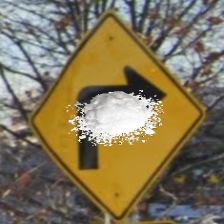}
        \caption{\footnotesize \centering Turn Right\newline Snowball 7}
        \label{fig:right_snowball_7}
    \end{subfigure}
    \begin{subfigure}[b]{0.101\textwidth}
        \centering
        \includegraphics[width=\linewidth]{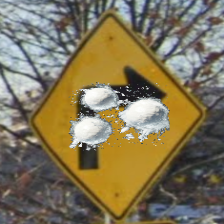}
        \caption{\footnotesize \centering Turn Right\newline Snowball 8}
        \label{fig:right_snowball_8}
    \end{subfigure}
    \begin{subfigure}[b]{0.1\textwidth}
        \centering
       \includegraphics[width=\linewidth]{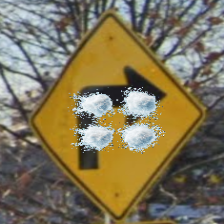}
        \caption{\footnotesize \centering Turn Right\newline Snowball 9}
        \label{fig:right_snowball_9}
    \end{subfigure}
\caption{ Turn Right Snowball Adversarial images.}
    \label{fig:right_snowball_adversarial_image}
\end{figure*}

The proposed search algorithm systematically identifies the optimal placement and orientation of the adversarial snowball overlay to maximize the likelihood of misclassification. Given a traffic sign image and its corresponding mask, the algorithm first determines the valid regions for perturbation by analyzing the non-zero pixels within the mask. The patch size is computed as a function of the available perturbable area, ensuring proportional scaling across different sign types. The algorithm then iterates over feasible placement positions, testing various patch rotations when applicable. For each candidate perturbation, an adversarial image is generated and evaluated using a pre-trained classifier. The confidence score of the incorrect prediction is recorded, and the perturbation configuration yielding the highest misclassification confidence is selected. This iterative search process ensures an efficient exploration of the perturbation space while optimizing adversarial effectiveness. Figures~\ref{fig:stop_snowball_adversarial_image} to~\ref{fig:right_snowball_adversarial_image} show the generated adversarial snowball images on each of the traffic signs.

\subsection{Optimized Search Algorithm}

The optimized search algorithm enhances the efficiency of adversarial perturbation generation by refining the search space iteratively. Initially, the algorithm determines valid perturbation regions using a binary mask and selects potential overlay positions. The snowball perturbation is applied across multiple patch sizes and orientations, with each configuration evaluated for its ability to induce misclassification. To improve search efficiency, the algorithm prioritizes high-confidence misclassifications and iteratively refines the mask by shrinking it around the most effective perturbation locations. This targeted reduction of the search space enables faster convergence towards an optimal adversarial configuration. Additionally, the use of structured patch scaling and selective angle testing reduces redundant computations while maintaining adversarial effectiveness. By iterating through progressively refined masks, the algorithm minimizes unnecessary evaluations and reduces the time to find best location for the snowball patch.

\section{Attack Results}
\label{attack_results}

This section presents the evaluation of the snowball adversarial attack.

\subsection{StreetView baseline Performance}

The evaluation is conducted on 5 randomly selected Street View images -- Stop, Yield, Pedestrian Crossing, Merge and Turn Right. LISA-CNN performs impressively, with correctly predicted images achieving an average confidence score exceeding 81\% for the five images although it was never trained on Street View images.
 The Stop sign had a confidence score of 85.42\%, the Yield sign had 68.30\%, the Pedestrian Crossing sign had 86.74\%, the Merge sign had 76.51\%, and the Turn Right sign had the highest confidence at 94.85\%. These scores provide a reference point for evaluating the effects of adversarial modifications.

\begin{table*}[t]
\centering
\caption{Accuracy (\%) of the adversarial images, representing the model's confidence in its predictions. In all but two cases, the attack has been successful and the accuracy reflects the confidence of LISA-CNN in the incorrectly predicted image.}

\begin{adjustbox}{width=0.98\textwidth}
\label{table_adversarial_confidence}
\small
\begin{tabular}{|p{1.5cm}|p{1.5cm}|p{1.5cm}|p{1.5cm}|p{1.5cm}|p{1.5cm}|p{1.5cm}|p{1.5cm}|p{1.5cm}|p{1.5cm}|}
\hline 
\textbf{Adversarial Image} & \textbf{Snowball 1}  & \textbf{Snowball 2} & \textbf{Snowball 3} & \textbf{Snowball 4} & \textbf{Snowball 5} & \textbf{Snowball 6} & \textbf{Snowball 7} & \textbf{Snowball 8} & \textbf{Snowball 9}
\\ \hline \hline
Stop & 76.61 & 69.64 & 33.40 & 59.99& 36.33 & 38.55  & 82.38 & 62.12 &52.91 \\ \hline 
Yield  & 49.87 & 44.14 & 25.50  & 41.15& 52.88& 55.26 &-& 32.15  & - \\ \hline 
Ped. Crossing  &91.41  &90.14 & 81.97 &89.36 &86.65 &84.90  &92.68  &93.73&88.54  \\ \hline 
Merge & 96.90 & 96.92 &96.41 &96.65  &96.88 &96.86 &96.79 &96.90 &96.86 \\ \hline 
Turn Right & 84.31 & 73.80& 42.25  &63.09 &57.49 &51.46 &77.08 &81.21 &80.27 \\ \hline  
\end{tabular}
\end{adjustbox}
\end{table*}

\begin{table*}[t]
\centering
\caption{Predicted labels of the adversarial images. In all but two cases the attack worked. The labels correspond to the confidences shown in entries in Table~\ref{table_adversarial_confidence}.}
\begin{adjustbox}{width=0.98\textwidth}
\label{table_adversarial_labels}
\small
\begin{tabular}{|p{1.5cm}|p{1.5cm}|p{1.5cm}|p{1.5cm}|p{1.5cm}|p{1.5cm}|p{1.5cm}|p{1.5cm}|p{1.5cm}|p{1.5cm}|}
\hline 
\textbf{Adversarial Image} & \textbf{Snowball 1}  & \textbf{Snowball 2} & \textbf{Snowball 3} & \textbf{Snowball 4} & \textbf{Snowball 5} & \textbf{Snowball 6} & \textbf{Snowball 7} & \textbf{Snowball 8} & \textbf{Snowball 9}
\\ \hline \hline
Stop & Speed Limit 25 & Yield& Speed Limit 25& Yield& Speed Limit 45& Signal Ahead& Speed Limit 25& Speed Limit 25& Speed Limit 25\\ \hline 
Yield  & Speed Limit 35 & Speed Limit 35& Speed Limit 35& Speed Limit 35& Ped. Crossing& Ped. Crossing& -& Speed Limit 35& - \\ \hline
Ped. Crossing  & Stop Ahead & Stop Ahead& Stop Ahead& Stop Ahead& Stop Ahead& Stop Ahead& Stop Ahead& Stop Ahead& Stop Ahead \\ \hline  
Merge & Ped. Crossing & Ped. Crossing & Ped. Crossing & Ped. Crossing & Ped. Crossing & Ped. Crossing & Ped. Crossing & Ped. Crossing & Ped. Crossing   \\ \hline 
Turn Right & Stop & Stop & Added Lane & Stop & Added Lane & Ped. Crossing & Stop & Stop & Stop \\ \hline  
\end{tabular}
\end{adjustbox}
\end{table*}

\subsection{Snowball Baseline Adversarial Attack}

Table~\ref{table_adversarial_confidence} shows the confidence scores for adversarial images generated by the Snowball Baseline Adversarial Attack, while Table~\ref{table_adversarial_labels} details their corresponding predicted labels across nine iterations. For instance, the ``Stop'' image exhibits considerable variability -- with predictions oscillating between ``Speed Limit 25'', ``Yield'', ``Speed Limit 45'', and ``Signal Ahead'' -- reflecting the fluctuating confidence scores that range from as low as 33.40\% to as high as 82.38\%. Similarly, the ``Yield'' image is mostly misclassified as ``Speed Limit 35'', though occasional outputs like ``Ped. Crossing'' and missing predictions hint at instability under adversarial conditions. In contrast, the ``Ped. Crossing'' image consistently yields ``Stop Ahead'' across all iterations, accompanied by high confidence levels (81.97\% to 93.73\%), indicating a systematic misclassification. The ``Merge'' image is uniformly labeled as ``Ped. Crossing'' with exceptionally high confidence (over 96\%), and the ``Turn Right'' image presents a mix of predictions -- primarily ``Stop'' with intermittent outputs like ``Added Lane'' and ``Ped. Crossing''. Overall, these variations not only demonstrate the Snowball attack's capability to induce misclassifications but also highlight the nuanced behavior of the model's decision boundaries in response to adversarial perturbations.

\subsection{Snowball Optimized Adversarial Attack}
The Snowball Optimized Adversarial Attack demonstrated varying degrees of performance improvement (i.e. lower or faster time is better) across different traffic signs at the 75\%, 50\%, and 25\% thresholds, as presented in Tables \ref{table_adversarial_timing_75}, \ref{table_adversarial_timing_50}, and~\ref{table_adversarial_timing_25}. 

At the 75\% threshold, the Stop sign shows a gradual decrease in timing, with the most significant reduction of 14\% in Snowball 9. Yield signs exhibit an initial decrease, followed by increases in timing from Snowball 2 to Snowball 7, with the highest positive change of 17\% in Snowball 2. Pedestrian Crossing signs consistently experience timing reductions, with the largest drop of 19\% in Snowball 9. The Merge and Turn Right signs also show reductions, with Merge timing decreasing by 20\% in Snowball 1 and Turn Right experiencing a peak decline of 21\% in Snowball 3.

\begin{table*}[th!]
\centering
\caption{Timing of finding the best location for the placement of snowballs on each of the images. Time units are seconds (s).}
\begin{adjustbox}{width=0.98\textwidth}
\label{table_adversarial_timing}
\small
\begin{tabular}{|p{1.5cm}|p{1.5cm}|p{1.5cm}|p{1.5cm}|p{1.5cm}|p{1.5cm}|p{1.5cm}|p{1.5cm}|p{1.5cm}|p{1.5cm}|}
\hline 
\textbf{Adversarial Image} & \textbf{Snowball 1}  & \textbf{Snowball 2} & \textbf{Snowball 3} & \textbf{Snowball 4} & \textbf{Snowball 5} & \textbf{Snowball 6} & \textbf{Snowball 7} & \textbf{Snowball 8} & \textbf{Snowball 9}
\\ \hline \hline
Stop & 5573  & 5608 & 5670 & 5671& 5716 & 5767 &5726&5743&5746 \\ \hline 
Yield  & 1010  &996 &991  & 1006& 987&1010 &1027 &1112  & 1125 \\ \hline 
Ped. Crossing  &3403  &3437 &3460 &3389 &3377 &3405&3443 &3709 &3635 \\ \hline 
Merge & 733 &757 &746 &745  &760 &770 &786 &718 &710 \\ \hline 
Turn Right & 1896 & 1897&1923  &1787 &1686 &1707 &1735 &1723 &1728 \\ \hline  
\end{tabular}
\end{adjustbox}
\end{table*}

\begin{table*}[th!]
\centering
\caption{Timing of finding the best location for the placement of snowballs on each of the images when 75\% of the mask area is used after finding the optimal point for the first snowball adversarial image. Time units are seconds (s). Lower value and negative \% are better.}
\begin{adjustbox}{width=0.98\textwidth}
\label{table_adversarial_timing_75}
\small
\begin{tabular}{|p{1.5cm}|p{1.5cm}|p{1.5cm}|p{1.5cm}|p{1.5cm}|p{1.5cm}|p{1.5cm}|p{1.5cm}|p{1.5cm}|p{1.5cm}|}
\hline 
\textbf{Adversarial Image} & \textbf{Snowball 1}  & \textbf{Snowball 2} & \textbf{Snowball 3} & \textbf{Snowball 4} & \textbf{Snowball 5} & \textbf{Snowball 6} & \textbf{Snowball 7} & \textbf{Snowball 8} & \textbf{Snowball 9}
\\ \hline \hline
Stop & 5630 \newline (+1\%)  & 5110 \newline  (-9\%) & 5052 \newline (-11\%) & 5087 \newline (-10\%)& 5063 \newline (-11\%)  & 5043 \newline (-13\%) &5236 \newline (-9\%)
&5000 \newline (-13\%)
&4953 \newline (-14\%) \\ \hline 
Yield  & 947 \newline (-6\%) &1169 \newline (+17\%)&1129 \newline (+14\%)  & 1175 \newline (+17\%)& 1152 \newline (+17\%) &1169 \newline (+16\%) &1182 \newline (+15\%) &1173 \newline (+6\%)  &  1155 \newline (+3\%)\\ \hline 
Ped. Crossing  &3230 \newline (-5\%) &2901 \newline (-16\%)
&2927 \newline (-15\%) &2893 \newline (-15\%)  &2954 \newline (-13\%) &2926 \newline (-14\%) &2951 \newline (-14\%)  &3046 \newline (-18\%) &2958 \newline (-19\%)  \\ \hline 
Merge 
&583 \newline (-20\%) &613 \newline (-19\%)  &636 \newline (-15\%)   &655 \newline (-12\%)  
&643 \newline (-15\%) 
&653 \newline (-15\%)  &661 \newline (-16\%) &653 \newline (-9\%)  &645 \newline (-9\%)  \\ \hline 
Turn Right 
&1757 \newline (-7\%)  &1573 \newline (-17\%) &1529 \newline (-21\%) &1587 \newline (-11\%) &1552 \newline (-8\%) &1566 \newline (-8\%) &1571 \newline (-9\%)  &1556 \newline (-10\%)  &1581 \newline (-9\%)  \\ \hline  
\end{tabular}
\end{adjustbox}
\end{table*}

\begin{table*}[th!]
\centering
\caption{Timing of finding the best location for the placement of snowballs on each of the images when 50\% of the mask area is used after finding the optimal point for the first snowball adversarial image. Time units are seconds (s). Lower value and negative \% are better.}
\begin{adjustbox}{width=0.98\textwidth}
\label{table_adversarial_timing_50}
\small
\begin{tabular}{|p{1.5cm}|p{1.5cm}|p{1.5cm}|p{1.5cm}|p{1.5cm}|p{1.5cm}|p{1.5cm}|p{1.5cm}|p{1.5cm}|p{1.5cm}|}
\hline 
\textbf{Adversarial Image} & \textbf{Snowball 1}  & \textbf{Snowball 2} & \textbf{Snowball 3} & \textbf{Snowball 4} & \textbf{Snowball 5} & \textbf{Snowball 6} & \textbf{Snowball 7} & \textbf{Snowball 8} & \textbf{Snowball 9}
\\ \hline \hline
Stop &5698 \newline (+2\%)  &3137 \newline (-44\%)  &3202 \newline (-44\%) 
&3141 \newline (-45\%)& 3183 \newline  (-44\%)  &3208 \newline (-44\%) & 3204 \newline  (-44\%)
& 3255 \newline (-44\%)
& 3224 \newline  (-44\%)\\ \hline 
Yield  &1035 \newline (+2\%)  
&911 \newline (-8\%) &907 \newline (-8\%) & 920 \newline  (-9\%)& 909 \newline (-8\%)
&905 \newline (-10\%) &913 \newline (-11\%) &910 \newline (-18\%)&  918 \newline  (-18\%)\\ \hline 
Ped. Crossing  &3287 \newline  (-3\%)
&1732 \newline (-50\%) &1743 \newline (-50\%) &1779 \newline (-48\%) &1704 \newline (-50\%) &1633 \newline (-52\%) &1648 \newline (-52\%)  &1775 \newline (-52\%) &1695 \newline (-53\%) \\ \hline 
Merge & 800 \newline (+9\%) &623 \newline (-18\%) &601 \newline (-20\%) &636 \newline (-15\%) &713 \newline (-6\%) &712 \newline (-8\%) &728 \newline (-7\%) &730 \newline (+2\%) &731 \newline (+3\%)\\ \hline 
Turn Right & 1794 \newline (-5\%) 
&685 \newline (-64\%) &642 \newline (-67\%) &630 \newline (-65\%) &606 \newline (-64\%) &622 \newline (-64\%) &621 \newline (-64\%) &643 \newline (-63\%) &635 \newline (-63\%)\\ \hline  
\end{tabular}
\end{adjustbox}
\end{table*}

\begin{table*}[th!]
\centering
\caption{Timing of finding the best location for the placement of snowballs on each of the images when 25\% of the mask area is used after finding the optimal point for the first snowball adversarial image. Time units are seconds (s). Lower value and negative \% are better.}
\begin{adjustbox}{width=0.98\textwidth}
\label{table_adversarial_timing_25}
\small
\begin{tabular}{|p{1.5cm}|p{1.5cm}|p{1.5cm}|p{1.5cm}|p{1.5cm}|p{1.5cm}|p{1.5cm}|p{1.5cm}|p{1.5cm}|p{1.5cm}|}
\hline 
\textbf{Adversarial Image} & \textbf{Snowball 1}  & \textbf{Snowball 2} & \textbf{Snowball 3} & \textbf{Snowball 4} & \textbf{Snowball 5} & \textbf{Snowball 6} & \textbf{Snowball 7} & \textbf{Snowball 8} & \textbf{Snowball 9}
\\ \hline \hline
Stop & 4937 \newline (-11\%)  
&757 \newline (-87\%) 
&751 \newline (-87\%) 
&776 \newline (-86\%) & 786 \newline  (-86\%)  &789 \newline (-86\%) & 794 \newline (-86\%) 
& 790 \newline (-86\%) 
& 791 \newline (-86\%)  \\ \hline 
Yield  & 1285 \newline (+27\%)   
&508 \newline (-49\%)  &474 \newline (-52\%)  & 419 \newline (-58\%) & 388 \newline  (-61\%) &383 \newline (-62\%) &371 \newline (-64\%) &369 \newline (-67\%)  & 394 \newline (-65\%) \\ \hline 
Ped. Crossing  &3184 \newline  (-6\%)  
&654 \newline (-81\%)  &665 \newline (-81\%) &677 \newline (-80\%) &638 \newline (-81\%) &606 \newline (-82\%)  &619 \newline (-82\%)  &615 \newline (-83\%) &589 \newline (-84\%)  \\ \hline 
Merge & 650 \newline (-11\%) 
&233 \newline (-69\%)   &248 \newline (-67\%) &240 \newline (-68\%) &236 \newline (-69\%) &246 \newline (-68\%) &251 \newline (-68\%) &248 \newline (-65\%) &242 \newline (-66\%)\\ \hline 
Turn Right & 1662 \newline (-12\%) 
&189 \newline (-91\%) &183 \newline (-90\%) &182 \newline (-90\%) &172 \newline (-90\%) &168 \newline  (-90\%) &176 \newline (-90\%) &176 \newline  (-90\%)  &180 \newline  (-90\%)\\ \hline  
\end{tabular}
\end{adjustbox}
\end{table*}

At the 50\% threshold, Stop signs timing initially increase by 2\% in Snowball 1, followed by a substantial decline, with the largest reduction of 45\% in Snowball 4. Yield signs show an initial increase of 2\% in timing, then decrease progressively, with the largest drop of 18\% in Snowball 9. Pedestrian Crossing signs experience severe improvements in timing, peaking at 53\% drop in Snowball 9, following a 3\% decrease in Snowball 1. Merge signs show a slight increase of 9\% in Snowball 1, followed by moderate reductions, with a recovery of a 3\% timing increase in Snowball 9. Turn Right signs experience extreme timing reductions, with the most significant drop of 67\% in Snowball 3, stabilizing around 63\% in later iterations.

At the 25\% threshold, Stop signs begin with a timing reduction of 11\% in Snowball 1, followed by a sharp decline, with the most notable reduction of 87\% in Snowball 3. Yield signs show an initial increase in time of 27\% in Snowball 1, but drop consistently in later iterations, with the largest decrease of 67\% in Snowball 8. Pedestrian Crossing signs experience a similar pattern, with an initial reduction of 6\%, followed by drastic decreases in time, culminating in a 84\% reduction in Snowball 9. Merge signs exhibit steady decreases, starting with 11\% in Snowball 1 and continuing to 66\% in Snowball 9. Turn Right signs show the most severe timing reductions, starting with a 12\% drop in Snowball 1 and reaching 91\% in Snowball 2, with only slight fluctuations thereafter, stabilizing at around 90\% in Snowball 9.

While signs such as Stop and Yield show moderate improvement in timing at the 75\% and 50\% thresholds, they exhibit more severe performance improvements at the 25\% threshold, particularly in later Snowball iterations. Pedestrian Crossing and Turn Right signs experience consistent and substantial reductions in timing across all thresholds, with the most extreme improvement observed at the 25\% threshold.
The findings underscore the increasing effectiveness of the attack as the threshold decreases, leading to progressively more significant timing improvement, especially in signs subjected to severe changes.

\section{Related Work}
\label{sec:related_work}

This section provides an overview of recent work on   adversarial example attacks on traffic sign classification.

 RP2 (Robust Physical Perturbations)~\cite{eykholt2018robust} was a pioneering approach that generated physically realizable adversarial examples capable of misleading deep neural networks under real-world conditions, including changes in lighting and angles. Unlike digital attacks, RP2 optimized perturbations that can be physically applied to traffic signs using carefully designed stickers, ensuring transferability across models while maintaining adversarial effectiveness in diverse environments. Subsequent works such as~\cite{liu2023adversarial} have extended RP2 by leveraging generative models to enhance robustness and stealth, further highlighting the security risks in deep learning-based traffic sign recognition systems.

 In~\cite{zhong2022shadows}, the authors introduced a novel optical adversarial attack that leveraged naturally occurring shadows to generate imperceptible yet effective perturbations. This approach exploited a common natural phenomenon to create stealthy adversarial examples in the physical world. The attack operated under a black-box setting and was evaluated extensively in both simulated and real-world environments, demonstrating its ability to mislead machine learning-based vision systems without drawing human attention. 

 The authors in~\cite{hsiao2024natural} explored the vulnerabilities of machine learning systems introduced by common illumination sources such as sunlight and flashlights. They proposed a novel attack method that simulated these lighting conditions to deceive machine learning models without requiring conspicuous physical modifications. Unlike traditional physical adversarial attacks, their approach employed a model-agnostic black-box attack, leveraging the Zeroth-Order Optimization (ZOO) algorithm~\cite{chen2017zoo} to identify deceptive patterns in a physically realizable space. They showcased the attack’s effectiveness, successfully misleading traffic sign classifiers in both digital and real-world settings.

 In~\cite{etim2024fallleafadversarialattack}, the authors introduced a new class of adversarial attacks that exploited fallen leaves to induce misclassification in street sign recognition systems. Unlike traditional adversarial attacks, these perturbations offer plausible deniability, as a leaf obstructing a sign could plausibly originate from a nearby tree rather than being deliberately placed by an attacker. Their results demonstrated the efficacy of naturally occurring perturbations in inducing misclassification, underscoring the potential to degrade the reliability of AI-driven traffic sign recognition systems in real-world 
 ~conditions.

\section{Conclusion}
\label{sec_conclusion}

This work introduced the Snowball Adversarial Attack, an iterative approach that progressively applies perturbations, referred to as ``snowballs,'' to traffic sign images. Unlike traditional adversarial attacks that optimize a single global perturbation, the Snowball Attack amplifies its effects incrementally over successive iterations, exploiting layer-wise vulnerabilities within the system. Experimental results demonstrate the effectiveness of this method in significantly degrading the performance of traffic sign recognition systems. At the 25\% threshold, the attack resulted in up to 90\% performance improvement for certain traffic signs, while the 50\% and 75\% thresholds showed substantial, though slightly less severe, performance improvements. These findings reveal the ability of the Snowball Attack to compound perturbations across layers, emphasizing the vulnerabilities of traffic sign recognition systems to such adversarial perturbations. Given the importance of reliable traffic sign recognition in autonomous driving and other safety-critical applications, these results highlight the need for stronger defenses against such adversarial strategies. Future work should focus on developing more robust models and effective defense mechanisms to mitigate the impact of compounded adversarial attacks.

\section*{Acknowledgements}

This work was supported in part by National Science Foundation grant \nsf{2245344}.
\bibliographystyle{ACM-Reference-Format}
\bibliography{bibtex/references}

\end{document}